\pdfoutput=1

\documentclass[11pt]{article}

\usepackage[final]{acl}
\usepackage{graphicx}
\usepackage{float}
\usepackage{times}
\usepackage{latexsym}
\usepackage{enumitem}
\usepackage[T1]{fontenc}

\usepackage[utf8]{inputenc}

\usepackage{microtype}

\usepackage{inconsolata}

\usepackage{graphicx}
\usepackage{amsmath}
\usepackage{amssymb}
\usepackage{amsfonts}
\usepackage{mathtools}

\usepackage{colortbl}
\definecolor{new_green}{rgb}{0.67, 0.88, 0.69}
\definecolor{max_green}{rgb}{0.31, 0.78, 0.47}

\title{CLaC at DISRPT 2025: Hierarchical Adapters for Cross-Framework Multi-lingual Discourse Relation Classification}

\author{Nawar Turk,  Daniele Comitogianni,   Leila Kosseim\\
  Computational Linguistics at Concordia (CLaC) Lab \\
   Dept. of Computer Science and Software Engineering\\
  Concordia University, Montréal, Québec, Canada \\
     \href{mailto:nawar.turk@mail.concordia.ca}{nawar.turk@mail.concordia.ca}, \href{mailto:danielecomitogianni@outlook.com}{danielecomitogianni@outlook.com}, \href{mailto:leila.kosseim@concordia.ca}{leila.kosseim@concordia.ca}
  }

\begin{document}
\maketitle

\begin{abstract}
We present our submission to Task 3 (Discourse Relation Classification) of the DISRPT 2025 shared task. Task 3 introduces a unified set of 17 discourse relation labels across 39 corpora in 16 languages and six discourse frameworks, posing significant multilingual and cross‑formalism challenges. We first benchmark the task by fine‑tuning multilingual BERT‑based models (\textit{mBERT}, \textit{XLM‑RoBERTa‑Base}, and \textit{XLM‑RoBERTa‑Large}) with two argument‑ordering strategies and progressive unfreezing ratios to establish strong baselines. We then evaluate prompt‑based large language models (namely Claude Opus 4.0) in zero‑shot and few‑shot settings to understand how LLMs respond to the newly proposed unified labels. Finally, we introduce HiDAC, a Hierarchical Dual‑Adapter Contrastive learning model. Results show that while larger transformer models achieve higher accuracy, the improvements are modest, and that unfreezing the top 75\% of encoder layers yields performance comparable to full fine‑tuning while training far fewer parameters. Prompt‑based models lag significantly behind fine‑tuned transformers, and HiDAC achieves the highest overall accuracy (67.5\%) while remaining more parameter‑efficient than full fine‑tuning.
\end{abstract}

\section{Introduction}
The 2025 DISRPT shared task addresses three challenges in discourse parsing\footnote{\url{https://sites.google.com/view/disrpt2025}}: discourse unit segmentation (Task~1), connective detection (Task~2), and discourse relation classification (Task~3) which aims at identifying logical and rhetorical relationships between text segments. 
This paper presents our approach to Task~3 which expands the task from previous years to 39 corpora across 16 languages and six discourse frameworks.

This year, the task proposed a unified set of 17 labels to classify the relations across multiple languages and frameworks, making the classification problem more challenging due to the diversity of data and the need to generalize across different frameworks and languages. This work contributes to the shared effort of building a multilingual and cross‑framework discourse parser. 

For our participation, we evaluated three methods: transformer-based baselines, prompt engineering with LLMs, and a custom made model HiDAC Hierarchical Dual-Adapter Contrastive learning, a novel parameter-efficient framework which employs a hierarchical adapter backbone and is trained with a dual-loss objective. Results show that HiDAC achieved slightly higher overall accuracy than the transformer baselines with considerably fewer training parameters, while the prompting approaches did not perform well.

\section{Related Work}
Recent work has explored cross-framework and multilingual discourse relation classification. \citet{costa2023mapping} proposed mappings between RST-DT and PDTB 3.0 frameworks, while \citet{costa2024multi, costa2025multi-lingual} developed multilabel hierarchical models for multilingual implicit discourse relation recognition. Recently, \citet{eichin-etal-2025-probing} conducted a comprehensive analysis evaluating open-weight LLMs, in which they defined a unified set of discourse relation labels to better understand how these models generalize across languages and annotation frameworks. The study demonstrated that LLMs, especially those with multilingual training corpora, can generalize discourse information across languages and frameworks. Their error analysis highlighted overlapping relations, such as frequent confusions between \textit{Elaboration}, \textit{Framing}, and \textit{Explanation}.

The DISRPT Task~3 itself was first introduced in 2021, where two teams submitted systems: DisCoDisCo~\citep{gessler-etal-2021-discodisco} achieved 61.82\%\footnote{\label{fn:disrpt2021}\url{https://sites.google.com/georgetown.edu/disrpt2021/results}} accuracy using language‑specific BERT-base models (varying per language) fine‑tuned per corpus with additional hand‑crafted features, while DiscRel~\citep{varachkina-pannach-2021-unified} reached 54.23\%\footref{fn:disrpt2021} accuracy using multilingual Sentence‑BERT to embed discourse units, then fed the Euclidean distance between the unit embeddings and directionality as features into a two-level stacked Random Forest, first predicting five coarse classes, then fine-grained labels within each class.

The task was proposed again in 2023 where three teams submitted systems: MELODI \citep{metheniti-etal-2023-discut} used three mBERT-based models: a baseline fine-tuned model and two adapter-enhanced variants with layer freezing, achieving 54.44\%\footref{fn:disrpt2023} accuracy. HITS~\citep{liu-etal-2023-hits} used BERT-based encoders with adversarial training, training separate models for large corpora and a joint multilingual model for smaller corpora and achieved an accuracy of 62.36\%\footref{fn:disrpt2023}. On the other hand, DiscoFlan~\citep{anuranjana-2023-discoflan}  used multilingual Flan-T5 based seq2seq model with instruction fine‑tuning to generate discourse relation labels from input prompts, achieving 31.21\%\footnote{\label{fn:disrpt2023}\url{https://sites.google.com/view/disrpt2023/results}} accuracy.

\section{Methodology}
\label{sec:method}
To address Task~3 (Discourse Relation Classification), we experimented with three families of models: (1)
BERT-based models as baselines (see Section~\ref{sec:bert});
(2) prompt-based generative models to use as a baselines as well (see Section~\ref{sec:prompt}), and
(3) a custom model, HiDAC, based on adapters and contrastive learning (see Section~\ref{sec:hidac}).

\subsection{Baseline Models}

\subsubsection{BERT-Based Models}
\label{sec:bert}
Our first baseline models are based on multilingual BERT transformers obtained from Hugging Face\footnote{ \url{https://huggingface.co/}}: \textit{bert-base-multilingual-cased}, \textit{xlm-roberta-base}, and \textit{xlm-roberta-large}. We conducted two sets of experiments: (1)  we tested both natural argument ordering as well as relation-directed ordering with no freezing of the BERT models; the relation ordering reorders the argument so as to respect the direction of the relation. For example, given the following instance with label \textit{purpose} and direction annotated as $1<2$:\\ 
\textit{Arg1: We propose a neural network approach} \\
\textit{Arg2: to benefit from the non-linearity of corpus-wide statistics for part-of-speech (POS) tagging.}
In this case, the natural ordering is the way the arguments appear in the text, whereas the relation-directed ordering is obtained by reversing the arguments to respect the direction of the relation $1<2$:\\
\textit{Arg1: to benefit from the non-linearity of corpus-wide statistics for part-of-speech (POS) tagging.} \\
\textit{Arg2: We propose a neural network approach}

(2) using the original order only, we explored gradual unfreezing strategies, where we initially froze all layers and then progressively unfroze 25\%, 50\%, and 75\% of the encoder top layers.

This multilingual encoder approach was tested because previous BERT-based work mainly fine-tuned single-language BERT-based models or combined small languages, with, to our knowledge, across both the 2021 and 2023 versions of the shared task~\citep{zeldes-etal-2021-disrpt,braud-etal-2023-disrpt}, only one team, MELODI \citep{metheniti-etal-2023-discut}, attempted a fully multilingual model. 
With DISRPT 2025's expanded data and languages, reevaluating a unified multilingual BERT‑based approach is now feasible and promising for cross‑lingual knowledge transfer.

\subsubsection{Prompt Engineering}
\label{sec:prompt}
As a second family of baselines, we evaluated both zero-shot and few-shot prompting strategies using Claude Opus 4.0. We chose the Claude model because in a previous work~\cite{Turk2025On}, this LLM achieved the highest average macroF1 compared to GPT and Gemini models that were evaluated to identify PDTB~3.0 Level~2 discourse relations.

Due to cost constraints of prompting the full development set, we used stratified sampling to ensure a representative evaluation. We first divided the validation set into 27 equal stratified subsets using \texttt{StratifiedKFold} from scikit-learn\footnote{\url{https://scikit-learn.org}} which preserves label distribution, we then randomly selected 4 groups ($\approx$1k samples each). Each selected subset was evaluated under both zero-shot and few-shot prompting settings.

In zero-shot prompting, we provided a structured prompt template (available in our repository\footnote{\url{https://github.com/CLaC-Lab/DISRPT-2025}}) without examples and tested both natural-order and relation-directed order argument arrangements. The template includes DISRPT 2025's unified set of 17 discourse relation labels in the label list.

For few-shot prompting, we used the same structured template but included examples. We developed a balanced pool of $\approx$1k examples, stratified across framework, language, and label (3 examples for each of the 327 unique framework-language-label combinations).
We then ran three few-shot prompting experiments: (1) Exp 1 used 4 examples in total (not per label), sampled randomly from the balanced example pool in the same language as the input instance (note that the task instructions remained in English). (2) Exp 2 used 4 English only examples from the same pool regardless of the instance language; and (3) Exp 3 analyzed Exp 1 \& 2 results to identify six labels with the lowest F1 scores, then used 6 English examples from these low-performing labels plus 2 random English examples from the remaining pool, doubling the number of examples.

\subsection{The HiDAC Model}
\label{sec:hidac}
Our third method which we call Hierarchical Dual-Adapter Contrastive (HiDAC) is based on a parameter-efficient fine-tuning approach. HiDAC is based on a pre-trained language model (PLM) with two main additions: (1) a hierarchical adapter backbone varying adapter types based on layer depth, and (2) a dual-loss training objective that applies two different losses at two different points in the network. An intermediate-layer contrastive loss is used to train the lower layers to build better representations, while a final-layer cross-entropy loss is used to train the upper layers for the classification task.

\subsubsection{Model Architecture}
\label{sec:archi}
\begin{figure}[t!]
    \centering
    \includegraphics[width=0.4\textwidth]{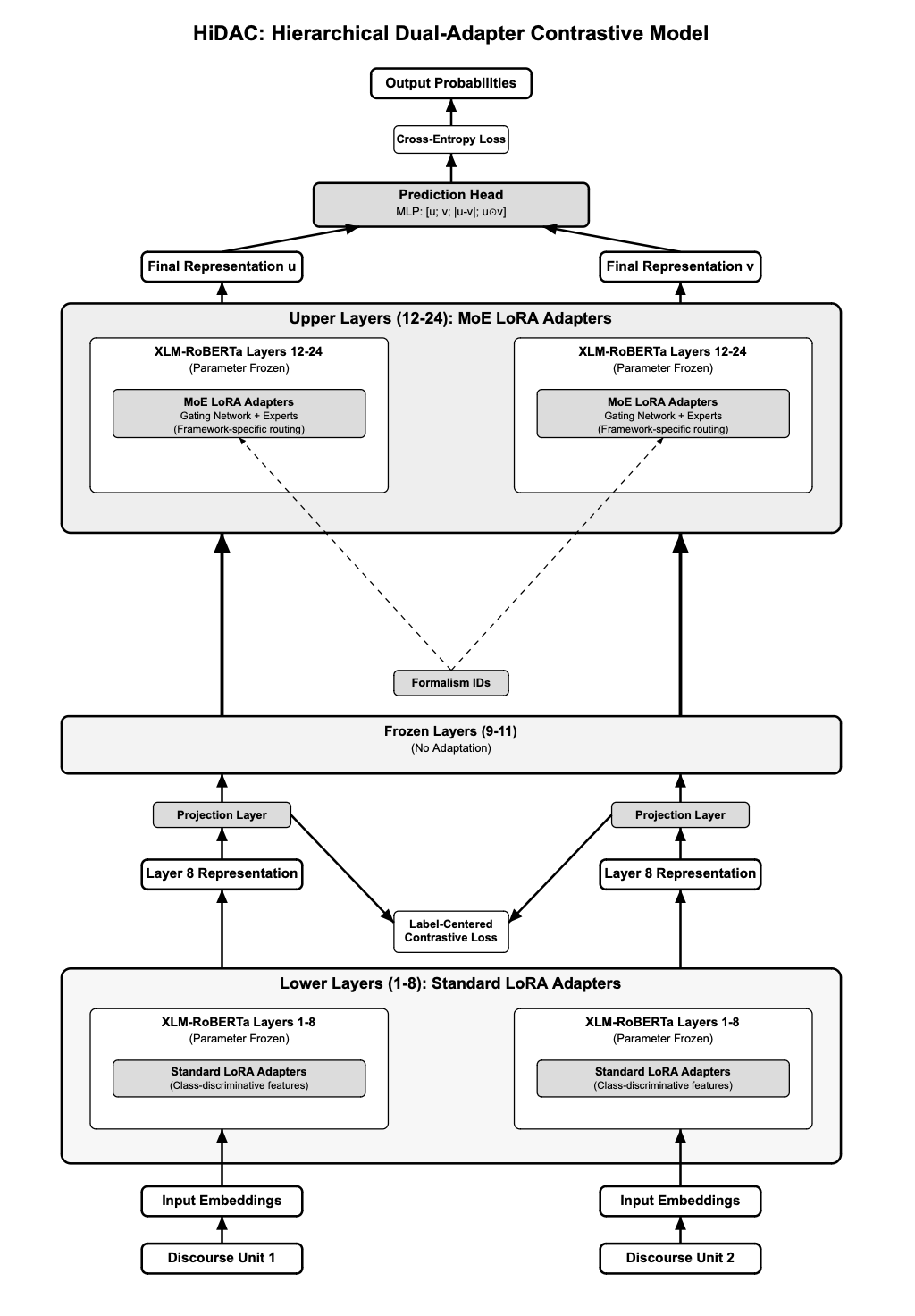}
    \caption{Overview of the HiDAC architecture. The contrastive loss is applied to an intermediate layer (8) to structure the embedding space, while the cross-entropy loss is applied to the representation at the last layer for the final classification.}
    \label{fig:hidac_architecture}
\end{figure}

The overall architecture of HiDAC is shown in Figure~\ref{fig:hidac_architecture}. The model processes each discourse argument independently through a dual-encoder PLM backbone \textit{(XLM-RoBERTa-Large)}. The architecture is named `hierarchical' because it partitions the encoder layers into two distinct functional levels, each with a different adaptation method and training objective. The lower layers (1-8) are trained using a Contrastive loss, which operates on the output representations from layer 8. The goal of this objective is to learn foundational representations by explicitly pulling the embedding of an instance closer to its correct class prototype, while simultaneously pushing it away from the prototypes of all incorrect classes. The upper layers (12-24) then receive these structured representations and are trained by a separate Cross-Entropy loss on the final-layer outputs to perform the classification task. The entire model is trained by minimizing a weighted sum of these two complementary losses.

\subsubsection{Discourse Unit Representation and Enhanced Prediction Head}
\label{sec:rep}
As shown in Figure~\ref{fig:hidac_architecture}, HiDAC uses a dual-encoder framework. From the final layer's hidden states, we extract the [CLS] token representation of each discourse unit, Arg1 and Arg2, resulting in vectors $\mathbf{u} \in \mathbb{R}^d$ and $\mathbf{v} \in \mathbb{R}^d$, respectively, where $d=1024$, the model size of \textit{XLM-RoBERTa-Large}.

To capture relational information for classification, we create an enhanced feature vector, $\mathbf{z}$, by concatenating $\mathbf{u}$ and $\mathbf{v}$ along with their element-wise difference and product \cite{Conneau2017Supervised}. We opted for this richer feature representation after an ablation study showed it outperformed the simpler concatenation of $[\mathbf{u}; \mathbf{v}]$, which achieved a development set accuracy of 67.18\% and an F1-score of 64.64\%. As this was lower than the 67.46\% accuracy and 65.14\% F1-score of the enhanced method (see Section~\ref{sec:results}), we chose the latter for our final model. The enhanced vector $\mathbf{z}$ serves as the input to the multi-layer perceptron (MLP) classifier and is defined as
$   \mathbf{z} = [\mathbf{u}~,~\mathbf{v}~,~\mathbf{u} - \mathbf{v}~,~\mathbf{u} * \mathbf{v}]
$.

\subsubsection{Hierarchical Adapter Architecture}
\label{sec:adapt}
HiDAC uses a hierarchical Parameter-Efficient Fine-Tuning (PEFT) strategy, segmenting the backbone PLM into lower layers (1 to 8) and upper layers (12 to 24), each employing distinct adapter mechanisms.

\paragraph{Lower Layers:} Layers (1-8) aim to learn foundational representations using standard LoRA adapters~\cite{Hu2021LoRA}. The purpose of these adapters is to learn basic patterns that help separate the different discourse relation classes. They achieve this by adjusting the model's early representations, grouping instances with the same discourse relation together in the embedding space. These adapters, optimized exclusively by the contrastive loss, add minimal computational overhead and encourage early formation of a well-organized embedding space.

\paragraph{Upper Layers:} Layers (12-24) aim to learn task-specific representations. The upper layers use Mixture-of-Experts LoRA adapters (MoE-LoRA), enabling dynamic, formalism-specific specialization. Each MoE-LoRA layer is made of multiple LoRA adapters acting as "experts." For each input token, a gating network computes a softmax distribution over these experts. The final output is a weighted sum of all expert outputs, allowing the model to learn a soft combination of specialized adaptations for each token. These adapters are optimized by the cross-entropy loss.

\subsubsection{Dual-Loss Objective} \label{sec:loss} HiDAC uses a dual-loss framework integrating Cross-Entropy loss ($L_{\mathrm{ce}}$) and Contrastive loss ($L_{\mathrm{lcl}}$) to optimize model representations and decision boundaries concurrently. The total loss ($L_{\mathrm{total}}$) is a weighted sum $ L_{\mathrm{total}} = \lambda_{\mathrm{ce}} \cdot L_{\mathrm{ce}} + \lambda_{\mathrm{cl}} \cdot L_{\mathrm{lcl}} $, where $\lambda_{\mathrm{ce}}$ and $\lambda_{\mathrm{cl}}$ are hyperparameters. \paragraph{Cross-Entropy Loss ($L_{\mathrm{ce}}$)} This loss is applied at the final classification layer as the primary objective for the prediction task. \paragraph{Contrastive Loss ($L_{\mathrm{lcl}}$)} This loss provides the training signal for the foundational adapters in the lower layers (1-8) of the encoder. We experimented with two contrastive objectives: a Label-Centered loss and a more traditional instance-vs-instance loss. 

As detailed in our analysis (see Section~\ref{sec:analysis}), the Label-Centered method yielded superior performance and training stability, and was therefore selected for our final model. The chosen Label-Centered SCL, inspired by~\cite{Wu2024ConstrainedMultiLayer}, simplifies the contrastive objective by introducing stable, learnable embeddings corresponding directly to each class label. Instead of comparing an instance to other random instances, the model's task is now much clearer: it learns to make the representation of an instance (from layer 8) more similar to the embedding of its correct label, while making it dissimilar to the embedding of all incorrect labels. The loss is computed as: 
$$L_{\mathrm{lcl}} = -\log \frac{\exp(\mathrm{sim}(\mathbf{h}_{\mathrm{contrastive}}, \mathbf{e}_y) / \tau)}{\sum_{j \in C} \exp(\mathrm{sim}(\mathbf{h}_{\mathrm{contrastive}}, \mathbf{e}_j) / \tau)}
$$

Here, $\mathbf{h}_{\mathrm{contrastive}}$ is formed by taking the [CLS] token representation for each discourse unit from the intermediate layer's output (layer~8), and then averaging these two vectors. The terms $\mathbf{e}_y$ and $\mathbf{e}_j$ refer to the learnable class prototypes; where $\mathbf{e}_y$ is the embedding of the ground-truth class $y$, while the sum in the denominator is over the embeddings for all 17 labels in the set $C$. Finally, $\mathrm{sim}$ is the cosine similarity, and $\tau$ is the temperature hyperparameter. 

On the other hand, the instance-vs-instance contrastive loss works by creating two slightly different vector representations for the same input text. The model is then trained to solve a simple matching task: it learns to pull these two views of the same text together, while simultaneously pushing them away from the representations of all other different texts in the batch. To provide a larger and more consistent set of negative examples, this method is augmented with a momentum-updated encoder and a negative queue~\cite{He2020Momentum}.

\section{Experimental Setup}
\subsection{Datasets}
\label{sec:data}

We trained and evaluated our models using the datasets provided by the shared task organizers.\footnote{\url{https://github.com/disrpt/sharedtask2025}}
In total, the benchmark is composed of $39$ datasets, covering $16$ languages and $6$ frameworks.
These datasets were obtained from the following corpora: 
the Czech RST Discourse Treebank 1.0 \cite{11234/1-5174},
the Potsdam Commentary Corpus \cite{StedeNeumann2014,bourgonje-stede-lrec2020},
the COVID-19 Discourse Dependency Treebank \cite{nishida-matsumoto-2022-domain},
the Discourse Dependency TreeBank for Scientific Abstracts \cite{yang-li-2018-scidtb,yi-etal-2021-unifying,cheng-li-2019-zero},
the Genre Tests for Linguistic Evaluation corpus \cite{aoyama-etal-2023-gentle},
the Georgetown University Multilayer corpus \cite{Zeldes2017},
the RST Discourse Treebank \cite{carlson-etal-2001-building},
the Science, Technology, and Society corpus  \cite{potter2008interactional},
the University of Potsdam Multilayer UNSC Corpus \cite{zaczynska-stede-2024-rhetorical},
the Minecraft Structured Dialogue Corpus \cite{thompson-etal-2024-discourse},
the Strategic Conversations corpus \cite{asher-etal-2016-discourse},
the Basque RST Treebank \cite{IruskietaAranzabeIlarrazaEtAl2013},
the Persian RST Corpus \cite{shahmohammadi2021persian},
the ANNOtation DIScursive corpus \cite{afantenos-etal-2012-empirical},
the SUMM-RE corpus \cite{hunter-etal-2024-meeting,hunter_summre_2025},
the Dutch Discourse Treebank \cite{redeker-etal-2012-multi},
the Polish Discourse Corpus \cite{ogr:etal:24,lrec:coling:24},
the Cross-document Structure Theory News Corpus \cite{CardosoMazieroRosarioCastroJorgeEtAl2011},
the Russian RST Treebank \cite{toldova-etal-2017-rhetorical},
the RST Spanish Treebank \cite{da-cunha-etal-2011-development},
the RST Spanish-Chinese Treebank \cite{cao-etal-2018-rst},
the Georgetown Chinese Discourse Treebank \cite{peng_gcdt_2022,peng_chinese_2022},
the DiscoNaija corpus \cite{scholman2025disconaija},
the Penn Discourse Treebank \cite{prasad-etal-2014-reflections,PDTB3-Annotation-Manual},
the TED-Multilingual Discourse Bank (English) \cite{zeyrek-etal-2018-multilingual,zeyrek2019ted},
the LUNA Corpus Discourse Data Set \cite{tonelli-etal-2010-annotation,RiccardiStepanovChowdhury2016},
the Portuguese Discourse Bank  \cite{CRPC-DB-Portuguese,genereux-etal-2012-introducing},
the Thai Discourse Treebank \cite{thai_prasertsom_2024},
the Turkish Discourse Bank \cite{zeyrek-webber-2008-discourse,zeyrek-kurfali-2017-tdb},
and
the Chinese Discourse Treebank \cite{Zhou2014}.

The dataset is divided into official train/dev/test splits; however, some corpora are included only as surprise or out-of-domain (OOD) evaluation sets and are not represented in the training data. Additionally, four corpora are fully masked (i.e., argument texts are replaced with underscores), and two are partially masked due to LDC subscription requirements.

For our experiments, we used the official train and dev splits to train and evaluate our models, excluding the instances requiring an LDC subscription.
As Table~\ref{tab:class-distribution} shows, the class distribution of the training dataset ($\approx$170K instances) and development dataset ($\approx$28K instances) is not balanced with  \textit{elaboration} and \textit{conjunction} being overrepresented while  \textit{alternation} and  \textit{reformulation} are severely underrepresented.

\begin{table}[t]
\centering
\resizebox{\linewidth}{!}{%
\begin{tabular}{|rlcc|}
\hline
&\textbf{Label} & \textbf{Train ($\approx$170k)} & \textbf{Dev ($\approx$28k)} \\
\hline
1&elaboration      & 19.9\% & 23.3\% \\
2&conjunction      & 17.5\% & 16.5\% \\
3&causal           & 11.7\% & 10.6\% \\
4&temporal         & 9.4\%  & 8.1\%  \\
5&query            & 6.5\%  & 5.1\%  \\
6&contrast         & 5.2\%  & 4.6\%  \\
7&concession       & 4.5\%  & 4.7\%  \\
8&comment          & 4.0\%  & 3.3\%  \\
9&explanation      & 3.7\%  & 3.9\%  \\
10&purpose          & 3.1\%  & 3.7\%  \\
11&condition        & 3.0\%  & 2.7\%  \\
12&attribution      & 3.0\%  & 3.2\%  \\
13&organization     & 2.9\%  & 3.5\%  \\
14&frame            & 2.3\%  & 2.9\%  \\
15&mode             & 1.4\%  & 2.0\%  \\
16&reformulation    & 1.2\%  & 1.4\%  \\
17&alternation      & 0.7\%  & 0.7\%  \\ 

\hline
\end{tabular}}
\caption{Class distribution  of discourse relation labels in the training and development sets.}
\label{tab:class-distribution}
\end{table}

\begin{table*}[htbp]
    \centering
\resizebox{\textwidth}{!}{%
\begin{tabular}{|l| r| cccccc| cccccc| cccccc |}

\hline
&  &
\multicolumn{18}{c|}{\textbf{BERT-based Models}} \\
\cline{3-20}
&  
& \multicolumn{6}{c|}{\textbf{mBERT}} 
& \multicolumn{6}{c|}{\textbf{XLM-R-Base}} 
& \multicolumn{6}{c|}{\textbf{XLM-R-Large}} \\

\cline{3-20}
\textbf{Category} 
& \textbf{Size (\%)}
& NO & RO & 0 & 0.25 & 0.5 & 0.75
& NO & RO & 0 &  0.25 & 0.5 & 0.75
& NO & RO & 0 &  0.25 & 0.5 & 0.75 \\ 
\hline
\textbf{Overall} & -- 
& 64.4\% & 63.5\% & 53.4\% & 64.0\% & 64.1\% & 64.3\% 
& 65.4\% & 64.6\% & 33.3\% & 63.8\% & 65.3\% & 65.1\% 
& 66.4\% & 65.8\% & 32.2\% & 65.9\% & 66.5\% & \cellcolor{new_green}{66.8\%} \\
\hline
\multicolumn{20}{|l|}{\textbf{Framework}} \\

pdtb & 27.0 
& 65.6\% & 65.1\% & 51.6\% & 65.0\% & 65.5\% & 65.3\%
& 67.3\% & 66.6\% & 25.5\% & 65.5\% & 67.1\% & 67.1\%
& 68.8\% & 67.7\% & 22.9\% & 67.8\% & \cellcolor{new_green}{68.8\%} & 68.5\% \\

rst & 26.5
& 57.8\% & 57.4\% & 47.8\% & 58.4\% & 58.4\% & 58.5\%
& 59.2\% & 58.3\% & 28.7\% & 57.2\% & 59.5\% & 59.0\%
& 60.2\% & \cellcolor{new_green}{60.5\%} & 27.6\% & 60.2\% & 59.8\% & 60.2\% \\

dep & 16.7
& 72.2\% & 70.3\% & 62.5\% & 70.6\% & 71.2\% & 71.7\%
& 72.0\% & 69.8\% & 46.4\% & 71.7\% & 70.1\% & 71.2\%
& 72.2\% & 70.2\% & 46.1\% & 70.6\% & \cellcolor{new_green}{72.4\%} & 72.1\% \\

sdrt & 14.5
& 75.3\% & 74.1\% & 64.4\% & 74.7\% & 73.9\% & 75.0\%
& 75.6\% & 75.4\% & 49.5\% & 74.4\% & 75.8\% & 75.0\%
& 76.2\% & 75.6\% & 51.2\% & 76.4\% & \cellcolor{max_green}{76.7\%} & \cellcolor{max_green}{76.7\%} \\

erst & 12.5
& 56.0\% & 55.6\% & 46.8\% & 56.2\% & 55.7\% & 56.3\%
& 56.7\% & 56.9\% & 24.2\% & 54.4\% & 57.6\% & 57.2\%
& 58.2\% & 58.9\% & 21.6\% & 58.4\% & 59.5\% & \cellcolor{max_green}{60.5\%} \\

iso & 2.8
& 47.2\% & 45.7\% & 40.7\% & 47.2\% & 48.2\% & 46.3\%
& 52.1\% & 50.7\% & 29.2\% & 49.2\% & 53.4\% & 50.8\%
& 55.3\% & 51.4\% & 31.4\% & 53.0\% & 52.6\% & \cellcolor{max_green}{56.7\%} \\

\hline
\multicolumn{20}{|l|}{\textbf{Language}} \\
eng & 51.0
& 67.0\% & 66.0\% & 57.7\% & 66.5\% & 66.3\% & 66.7\%
& 67.6\% & 66.6\% & 38.0\% & 66.1\% & 67.2\% & 67.6\%
& 68.1\% & 67.2\% & 37.1\% & 67.7\% & 68.6\% & \cellcolor{new_green}{68.9\%} \\

zho & 9.8
& 60.3\% & 62.0\% & 47.6\% & 59.9\% & 60.4\% & 60.3\%
& 60.7\% & 61.8\% & 26.9\% & 60.1\% & 62.2\% & 60.6\%
& \cellcolor{new_green}{64.2\%} & 64.0\% & 24.9\% & 61.0\% & 63.9\% & 63.1\% \\

rus & 8.2
& 62.5\% & 61.2\% & 52.2\% & 62.0\% & 62.8\% & 62.9\%
& 63.8\% & 62.6\% & 30.4\% & 60.5\% & 63.5\% & 63.4\%
& 63.5\% & 63.8\% & 28.9\% & 63.7\% & 63.7\% & \cellcolor{new_green}{64.2\%} \\

por & 7.4
& 64.7\% & 64.2\% & 55.6\% & 65.8\% & 65.8\% & 65.2\%
& 67.0\% & 66.9\% & 33.3\% & 65.6\% & 66.9\% & 66.3\%
& 68.3\% & 68.0\% & 30.0\% & \cellcolor{new_green}{68.9\%} & 67.8\% & 68.0\% \\

tha & 4.5
& 93.5\% & 94.2\% & 66.3\% & 94.4\% & 94.0\% & 94.4\%
& 96.0\% & 95.9\% & 26.7\% & 95.0\% & \cellcolor{new_green}{96.1\%} & 95.7\%
& 95.7\% & 95.9\% & 26.5\% & 95.5\% & 95.9\% & 96.0\% \\

pcm & 3.8
& 51.0\% & 49.6\% & 33.3\% & 49.6\% & 48.8\% & 50.0\%
& 52.9\% & 51.0\% & 22.9\% & 47.4\% & 50.5\% & 51.7\%
& \cellcolor{new_green}{56.1\%} & 54.2\% & 21.0\% & 52.3\% & 53.5\% & 52.8\% \\

pol & 2.8
& 47.2\% & 45.7\% & 40.7\% & 47.2\% & 48.2\% & 46.3\%
& 52.1\% & 50.7\% & 29.2\% & 49.2\% & 53.4\% & 50.8\%
& 55.3\% & 51.4\% & 31.4\% & 53.0\% & 52.6\% & \cellcolor{max_green}{56.7\%} \\

eus & 2.2
& 50.3\% & 48.4\% & 37.8\% & 49.3\% & 50.8\% & 52.6\%
& 52.4\% & 51.3\% & 22.5\% & 52.3\% & 54.1\% & 54.2\%
& 54.6\% & 57.2\% & 21.8\% & 56.7\% & 53.7\% & \cellcolor{max_green}{58.6\%} \\

fra & 1.9
& 61.2\% & 54.5\% & 51.8\% & 60.6\% & 59.5\% & 59.1\%
& 58.5\% & 57.6\% & 38.4\% & 60.0\% & 59.3\% & 54.5\%
& 62.0\% & 58.1\% & 37.1\% & \cellcolor{max_green}{62.9\%} & 62.5\% & 62.5\% \\

fas & 1.8
& 55.1\% & 53.5\% & 45.9\% & 52.9\% & 53.5\% & 54.5\%
& 54.3\% & 54.3\% & 28.1\% & 54.7\% & 54.1\% & 55.5\%
& \cellcolor{max_green}{57.9\%} & 57.7\% & 30.1\% & 56.5\% & 56.5\% & 57.7\% \\

spa & 1.7
& 68.1\% & 65.6\% & 50.9\% & 67.7\% & 67.7\% & 69.2\%
& 66.2\% & 66.7\% & 27.5\% & 63.5\% & 68.3\% & 64.8\%
& 68.3\% & 68.1\% & 26.8\% & 68.3\% & \cellcolor{max_green}{70.4\%} & 66.9\% \\

deu & 1.6
& 44.7\% & 44.2\% & 37.6\% & 45.4\% & 47.6\% & 45.8\%
& 47.1\% & 46.2\% & 18.6\% & 47.3\% & 48.2\% & 45.4\%
& 49.3\% & 46.9\% & 19.2\% & \cellcolor{new_green}{49.3\%} & 45.6\% & 44.9\% \\

nld & 1.2
& 53.2\% & 51.1\% & 38.4\% & 54.4\% & 54.7\% & 56.5\%
& 56.5\% & 50.5\% & 29.9\% & 56.5\% & 54.7\% & 55.9\%
& 55.3\% & 57.1\% & 28.4\% & \cellcolor{max_green}{58.9\%} & 55.9\% & 54.7\% \\

tur & 0.8
& 51.7\% & 44.5\% & 44.1\% & 46.4\% & 50.2\% & 49.8\%
& 54.0\% & 45.0\% & 30.8\% & 52.1\% & 52.1\% & 52.6\%
& 52.6\% & 50.7\% & 32.2\% & 53.6\% & \cellcolor{max_green}{54.5\%} & \cellcolor{max_green}{54.5\%} \\

ita & 0.7
& 60.7\% & 63.1\% & 54.4\% & 62.1\% & 60.2\% & 59.2\%
& 63.1\% & 63.6\% & 22.8\% & 59.7\% & 59.2\% & 60.7\%
& 58.7\% & 60.7\% & 21.8\% & 62.6\% & 62.1\% & \cellcolor{max_green}{65.0\%} \\

ces & 0.4
& 40.7\% & 47.2\% & 33.3\% & 45.5\% & 39.8\% & 45.5\%
& 48.0\% & 46.3\% & 22.0\% & 41.5\% & 39.8\% & 44.7\%
& 45.5\% & \cellcolor{max_green}{55.3\%} & 23.6\% & 46.3\% & 53.7\% & 48.0\% \\
\hline
\end{tabular}}
\caption{Accuracy of the BERT-based models (\textit{mBERT}, \textit{XLM-RoBERTa-Base} \& \textit{XLM-RoBERTa-Large}) across the two argument ordering and freezing strategies with subset size (\%) on the development ($\approx$28k) dataset. Note: NO = Natural ordering; RO = Relation-based ordering. The percentages represent the portion of the encoder's top layers that we unfreeze after initially freezing the entire model. All progressive unfreezing experiments were conducted with NO argument ordering only. Light green cells indicate the highest accuracy within each model family for a given framework/language, while dark green cells indicate the overall best accuracy across all families.}
\label{tab:bert-models-size}
\end{table*}

\begin{table*}
\resizebox{\textwidth}{!}{%
\begin{tabular}{|l|r|cc|ccc|| ccc|}
\hline
&  
& \multicolumn{5}{c||}{\textbf{Prompt-based Models}} 
& \multicolumn{3}{c|}{\textbf{Custom Model}}
\\
\cline{3-10}
&   
& \multicolumn{2}{c|}{\textbf{Zero-Shot}} 
& \multicolumn{3}{c||}{\textbf{Few-Shot}} 
& \multicolumn{3}{c|}{\textbf{HiDAC}} \\
\cline{3-10}
\textbf{Category} 
& \textbf{Size (\%)} 
& NO & RO 
& Exp 1 &  Exp 2 & Exp 3 
& Instance * Instance & Label Centered & Final HiDAC \\
\hline
\textbf{Overall} & --   
& \cellcolor{new_green}{40.85\%} & 31.93\% & 39.90\% & 39.6\% & 34.7\%  & 66.88\% & 66.97\% & \cellcolor{max_green}{67.46\%} \\
\hline
\multicolumn{10}{|l|}{\textbf{Framework}} \\
pdtb & 27.0 
& 43.22\% & 35.30\% & \cellcolor{new_green}{43.50\%} & 41.8\% & 37.50\% & 67.89\% & 67.64\% & \cellcolor{max_green}{69.00\%} \\
rst  & 26.5 
& \cellcolor{new_green}{39.36\%} & 28.80\% & 36.33\%  & 36.9\% & 31.60\% & 60.63\% & \cellcolor{max_green}{60.98\%} & 60.92\% \\
dep  & 16.7 
& \cellcolor{new_green}{51.78\%} & 35.70\% & 48.60\% & 48.0\% & 38.65\%  & 75.70\% & \cellcolor{max_green}{75.96\%} & 75.89\% \\
sdrt & 14.5 
& 30.76\% & 31.30\% & 31.48\%  & \cellcolor{new_green}{32.2\%} & 31.13\%  & 76.17\% & 76.62\% & \cellcolor{max_green}{76.67\%} \\
erst & 12.5 
& 36.90\% & 26.05\% & 37.85\% & \cellcolor{new_green}{37.9\%} & 33.50\%  & 58.97\% & 58.68\% & \cellcolor{new_green}{59.49\%} \\
iso  & 2.8  
& 37.33\% & 39.18\% & 39.35\%  & \cellcolor{new_green}{40.4\%} & 36.55\%  & 50.79\% & 50.53\% & \cellcolor{new_green}{51.97\%} \\
\hline
\multicolumn{10}{|l|}{\textbf{Language}} \\
eng & 51.0  
& \cellcolor{new_green}{40.36\%} & 30.38\% & 39.90\%  & 40.0\% & 34.70\% & 69.67\% & 69.75\% & \cellcolor{max_green}{70.00\%} \\
zho & 9.8  
& 36.15\% & 27.68\% & \cellcolor{new_green}{37.18\%} & 35.5\% & 32.83\%  & 64.35\% & 64.24\% & \cellcolor{max_green}{65.06\%} \\
rus & 8.2  
& \cellcolor{new_green}{45.12\%} & 34.90\% & 39.15\% & 40.3\% & 37.53\%  & 64.17\% & 63.64\% & \cellcolor{max_green}{64.61\%} \\
por & 7.4  
& \cellcolor{new_green}{47.38\%} & 31.38\% & 41.53\% & 39.7\%  & 34.78\%  & 68.16\% & 68.36\% & \cellcolor{max_green}{69.92\%} \\
tha & 4.5  
& 64.30\% & 64.30\% & \cellcolor{new_green}{66.18\%} & 64.4\%& 57.70\%   & 95.82\% & 95.58\% & \cellcolor{max_green}{95.90\%} \\
pcm & 3.8  
& 28.23\% & 23.75\% & \cellcolor{new_green}{36.23\%} & 32.3\%  & 24.33\%  & 51.24\% & 50.76\% & \cellcolor{max_green}{53.33\%} \\
pol & 2.8  
& 37.33\% & 39.18\% &  39.35\%  & \cellcolor{new_green}{40.4\%} & 36.55\%  & 50.79\% & 50.53\% & \cellcolor{new_green}{51.97\%} \\
eus & 2.2  
& 33.18\% & 26.35\% & 34.83\% & \cellcolor{new_green}{35.9\%}  & 30.78\% & 55.54\% & \cellcolor{new_green}{55.70\%} & 53.75\% \\
fra & 1.9  
& \cellcolor{new_green}{45.63\%} & \cellcolor{new_green}{45.63\%} & 35.30\%  & 38.7\% & 37.38\%  & 56.02\% & 57.36\% & \cellcolor{new_green}{57.36\%} \\
fas & 1.8  
& 35.83\% & 29.60\% & 37.28\% & \cellcolor{new_green}{38.8\%} & 32.60\%  & 52.30\% & \cellcolor{new_green}{55.51\%} & 55.31\% \\
spa & 1.7  
& 34.68\% & 17.68\% & \cellcolor{new_green}{38.80\%} & 38.1\%  & 30.33\% & 66.88\% & 66.88\% &\cellcolor{new_green}{67.71\%} \\
deu & 1.6  
& \cellcolor{new_green}{27.25\%} & 25.88\% & 23.00\%  & 23.0\% & 23.95\%  & 56.86\% & \cellcolor{max_green}{57.52\%} & 56.64\% \\
nld & 1.2  
& \cellcolor{new_green}{34.50\%} & 34.18\% & 31.65\%  & 25.4\%  & 24.03\% & 49.85\% & \cellcolor{new_green}{53.47\%} & 52.27\% \\
tur & 0.8  
& \cellcolor{new_green}{40.08\%} & 28.98\% & \cellcolor{new_green}{40.08\%} & 36.5\%  & 19.05\% & 48.82\% & 51.66\% & \cellcolor{new_green}{52.61\%} \\
ita & 0.7  
& 33.75\% & 20.00\% & 30.00\%  & 30.0\% & \cellcolor{new_green}{37.50\%}  & \cellcolor{new_green}{60.68\%} & 60.19\% & 59.71\% \\
ces & 0.4  
& \cellcolor{new_green}{16.08\%} & \cellcolor{new_green}{16.08\%} & 6.25\%  & 13.4\% & 9.38\% & 49.59\% & \cellcolor{new_green}{51.22\%} & 49.59\% \\
\hline

\end{tabular}}
\caption{Accuracy of Prompt Engineering using Claude Opus 4.0 (Zero-Shot, Few-Shot) and HiDAC on the development ($\approx$28k) dataset. 
Note: NO = Natural ordering; RO = Relation-based ordering; 
Exp 1 = Multilingual examples; Exp 2 = English-only examples; Exp 3 = English-only mainly from weak-label examples. Light green cells indicate the highest accuracy within each model family for a given framework/language, while dark green cells indicate the overall best accuracy across all families.}
\label{tab:prompt-engineering-and-HiDAC}

\end{table*}


\subsection{Training Details}
\label{sec:training}
For the BERT-based models (see Section~\ref{sec:bert}), we used standard fixed hyperparameters: We use the AdamW optimizer with a learning rate of 3e-5, batch size of 32, weight decay of 0.01, and warm-up ratio of  0.1, we set a maximum of 20 epochs and applied early stopping with a patience of 3 epochs based on the validation loss and we used fixed random seeds (seed=42) for reproducibility across all experiments. We used the default \texttt{AutoModelForSequenceClassification} implementation from Hugging Face Transformers\footnote{\url{https://huggingface.co/docs/transformers}} with 17 output classes corresponding to DISRPT 2025’s unified label set.

For the prompt-based experiments (see Section~\ref{sec:prompt}), we set the temperature to 0 to ensure fully deterministic outputs. We also fixed a random seed (seed=42) when selecting  examples for the few-shot prompting setting for reproducibility.

The implementation of HiDAC uses \textit{XLM-RoBERTa-Large} as the foundational model.  Adapter modules are applied using forward hooks, preserving the pre-trained weights while updating only the small adapter matrices. During inference, the trained LoRA adapter weights are loaded on top of the frozen base model. This approach adds negligible computational latency compared to using the base model without any adapters, as the forward pass only involves a few small matrix multiplications~\cite{Hu2021LoRA}.
The hyperparameters were tuned empirically based on performance on the development set. Our final model uses a LoRA rank ($r$) of 128 with a scaling factor ($\alpha$) of 256 for all adapters. For the contrastive loss, the temperature ($\tau$) was set to 0.1. The final loss weighting coefficients were set to $\lambda_{\mathrm{ce}}=1.0$ and $\lambda_{\mathrm{cl}}=0.3$. We trained the model using the AdamW optimizer with a learning rate of 2e-5 and a cosine learning rate scheduler with a warmup period covering the first 2 epochs. The model was trained with a batch size of 32, fixed random seed (seed = 42) and we used early stopping with patience of 2 epochs based on the accuracy of the development set. All experiments were conducted on Google Colab using a single NVIDIA A100 GPU. Training run took approximately 5 hours for the argument-ordering and HiDAC models, while the progressive unfreezing ones were shorter and varied depending on the unfreezing ratio.

\section{Results}
\label{sec:results}
Tables~\ref{tab:bert-models-size} and~\ref{tab:prompt-engineering-and-HiDAC} summarize the results across all three model families where we report the accuracy values overall, as well as by framework and language. 
Table~\ref{tab:bert-models-size} shows the results for the two argument-order arrangements and for the progressive unfreezing experiments at 0\%, 25\%, 50\%, and 75\% unfreezing ratios. Overall, accuracy increases across unfreezing ratios for all models, with the highest scores observed at the 75\% unfreezing ratio, and natural argument ordering consistently outperformed relation-directed ordering across all models. 
Table~\ref{tab:prompt-engineering-and-HiDAC} shows prompting results, with zero-shot achieving 40.85\% and few-shot reaching 39.90\%.
HiDAC achieved the highest overall performance at 67.46\%, outperforming both the best BERT baseline without progressive unfreezing, where the model is fully trainable (\textit{XLM-RoBERTa-Large} at 66.43\%) and with progressive unfreezing (66.76\% with 75\% unfreezing). 

\section{Analysis}
\label{sec:analysis}

We analyze the results of Tables~\ref{tab:bert-models-size} and~\ref{tab:prompt-engineering-and-HiDAC} to better understand the impact of model choices.

\subsection{BERT-Based Models}

\paragraph{Effect of Model Size:}
As shown in Table~\ref{tab:bert-models-size} larger models tend to perform slightly better. \textit{XLM-RoBERTa-Large} (550M parameters) achieves 66.4\%, compared to \textit{XLM-RoBERTa-Base} (65.4\%, 270M parameters) and \textit{mBERT} (64.4\%, 179M parameters). Although the absolute gains are modest, this trend suggests that increased capacity may help the model capture complex semantic relationships across languages and annotation frameworks.

\paragraph{Effect of Progressive Unfreezing:}
Progressive unfreezing leads to modest performance improvements as more layers are unfrozen; accuracy steadily increases from the 25\% to the 75\% unfreezing ratio across all models. 
At 75\% unfreezing, performance is comparable to or slightly higher than full fine-tuning while being significantly more parameter-efficient. For example, \textit{XLM-RoBERTa-Large} achieves 66.8\% accuracy while training only 41\% of parameters (vs. 66.4\% fully trained), and \textit{mBERT} reaches 64.3\% while training 36\% of parameters (as opposed to 64.4\% fully trained). This suggests that full fine-tuning may be unnecessary to achieve strong performance, allowing reductions in computational resources.

\paragraph{Cross-Formalism Analysis:}
The training data is dominated by PDTB (28.3\%) and RST (31.8\%). In contrast, SDRT and DEP represent only 18.7\% and 4\% of the training data, respectively. However, on the development set, SDRT (14.5\% of dev) and DEP (16.7\% of dev) achieve notably higher accuracy than PDTB (27\% of dev) and RST (26.5\% of dev) across models. 
For example, on the XLM-R-Large with unfreezing ratio of 0.75, we have the following accuracy values: SDRT = 76.7\%, DEP = 72.1\% vs. PDTB = 68.5\%, RST = 60.2\%. 
This raises the question of whether the 17 proposed unified labels may align more with these frameworks or whether other factors, such as the ratio of implicit to explicit relations, contribute to this discrepancy. Further analysis is needed to understand the source of these differences.

\subsection{Prompt-based Models}
Prompt engineering was significantly affected by relation-directed ordering, suggesting that LLMs may prefer the natural flow of arguments regardless of discourse relation source or direction. Surprisingly, performance did not improve when using few-shot prompts with four examples, whether English-only or instance-language-specific, showing the difficulty of designing effective prompts for the proposed labels; particularly, for the following labels, which achieved F1 scores below 15\%: \textit{concession}, \textit{explanation}, \textit{frame}, \textit{mode}, \textit{organization}, and \textit{reformulation}. When doubling the examples in Exp 3, accuracy dropped to 34.7\%, which could be because the model became overwhelmed by conflicting patterns and noise. Too many examples could have created confusion rather than clarity.
The instructions in all cases were in English only; it is worth investigating the use of language-specific instructions. Also, we only investigated with the Claude model;  it may be beneficial to evaluate other LLMs on the unified discourse labels for a broader comparison.

\subsection{HiDAC Model}
\paragraph{Effect of Contrastive Learning Objective:}
We assess two distinct supervised contrastive learning (SCL) objectives.
As shown in Table~\ref{tab:prompt-engineering-and-HiDAC}, the instance-based supervised contrastive learning objective, augmented with a momentum encoder and negative queue~\cite{He2020Momentum} achieved strong performance, comparable to our other methods. However, a closer analysis of the training dynamics revealed a significant issue: the contrastive loss value remained stagnant after the initial warm-up period, indicating that the model was failing to optimize this objective. We hypothesize this was caused by two factors. First, random in-batch negatives often included semantically similar pairs (e.g., two \textit{elaboration} instances), which may have provided a contradictory training signal. Second, the constantly changing nature of the negative examples prevented the model from learning against a fixed target. Although the loss stagnated, it did lead to a strong performance.

The Label-Centered SCL improved the learning dynamics. Unlike the stagnant loss observed with the instance-vs-instance method, the Label-Centered objective resulted in a consistent decrease in the contrastive loss throughout training. This indicates that the model was able to effectively learn from the clearer, more stable training signal provided by the fixed label embeddings. While the final performance scores were comparable to the instance-based method, we selected the Label-Centered approach for our final model due to its demonstrably superior training stability.

\paragraph{HiDAC Final Tuning:}
Our final model incorporates few tuning enhancements  to the Label-Centered architecture. 

As a regularization technique, we used label smoothing within the cross-entropy loss function and applied gradient clipping during training. These methods mitigate overfitting, stabilize training dynamics, and facilitate smoother convergence.

In addition, we replaced the linear learning rate scheduler with a cosine annealing scheduler. This modification ensures a more gradual and stable reduction in learning rate, improving convergence behavior during the final training stages.

As shown in Table~\ref{tab:prompt-engineering-and-HiDAC}, these refinements delivered a final incremental performance boost, ultimately yielding our best overall results.

\section{Conclusion}
This paper presented three approaches for Task~3 of DISRPT 2025: transformer-based baselines, prompt-based models, and HiDAC, a hierarchical adapter-based model with a dual-loss objective. Our experiments showed that natural argument ordering and progressive unfreezing generally improved performance, while prompt-based approaches underperformed compared to fine-tuned transformers. HiDAC achieved the best overall accuracy while training on fewer parameters, showing that adapter-based methods can reduce training cost without sacrificing accuracy. Future work could focus on improving prompt-based models and exploring multi-task training across Tasks 1-3. Also we plan to explore a multi-layer contrastive loss, applying the SCL objective across several intermediate layers to build a more robust representation. Additionally, we will investigate using Focal Loss with class-aware weighting to better address the severe class imbalance in the dataset and improve performance on underrepresented labels.

\section{Limitations}

For the BERT-based models, we used fixed hyperparameters across all models and datasets without additional fine-tuning on the validation sets. Our prompt engineering experiments, although one setup included language-based examples, used instructions only in English. While we explored both zero-shot and few-shot prompting, we did not experiment with chain-of-thought reasoning and only used the Claude model.

With respect to HiDAC, the hierarchical dual-adapter design introduces additional computational overhead during training; specifically, calculating two separate losses and extracting representations from intermediate layers makes the training process slower than a standard approach. Furthermore,  experiments revealed that the model's performance is sensitive to the balance between the cross-entropy and contrastive loss weights ($\lambda_{\mathrm{ce}}$ and $\lambda_{\mathrm{cl}}$). This suggests that applying this framework to new datasets would require careful hyperparameter tuning.

The architectural choices of HiDAC could also be explored. The division of the PLM into lower (1-8) and upper (12-24) layers was based on prior work~\cite{Wu2024ConstrainedMultiLayer} and was not optimized; other partitioning schemes might yield different outcomes. Similarly, the gating mechanism in the upper-layer MoE adapters uses a simple soft mixture, and more advanced sparse routing strategies were not explored. Finally, the performance of the model is fundamentally dependent on the capabilities of the underlying XLM-RoBERTa backbone, and its effectiveness may vary when applied to different pre-trained encoders. 

Finally, due to time constraints, we did not perform tests of statistical significance on the performance difference of the various experiments.

\section*{Acknowledgments}
This work was financially supported by the Natural Sciences and Engineering Research Council of Canada (NSERC), the Fonds de recherche du Québec (FRQ), and the Pierre Arbour Foundation.

\bibliography{references}
\end{document}